\documentclass[sigconf, authorversion]{acmart}
\usepackage{graphicx}
\usepackage{hyperref} 
\usepackage{multirow}
\usepackage{arydshln} 
\usepackage[flushleft]{threeparttable} 
\usepackage{enumitem}

\AtBeginDocument{%
  \providecommand\BibTeX{{%
    \normalfont B\kern-0.5em{\scshape i\kern-0.25em b}\kern-0.8em\TeX}}}

\copyrightyear{2023} 
\acmYear{2023} 
\setcopyright{acmlicensed}
\acmConference[MM '23]{Proceedings of the 31st ACM International Conference on Multimedia}{October 29-November 3, 2023}{Ottawa, ON, Canada}
\acmBooktitle{Proceedings of the 31st ACM International Conference on Multimedia (MM '23), October 29-November 3, 2023, Ottawa, ON, Canada}
\acmPrice{15.00}
\acmDOI{10.1145/3581783.3613806}
\acmISBN{979-8-4007-0108-5/23/10}

\begin{document}

\title{SGDiff: A Style Guided Diffusion Model for Fashion Synthesis}

\author{Zhengwentai Sun}
\affiliation{%
  \institution{The Hong Kong Polytechnic University}
  \city{Hong Kong}
  \country{Hong Kong}
}
\email{zhengwt.sun@connect.polyu.hk}
\orcid{0000-0002-3884-137X}

\author{Yanghong Zhou}
\affiliation{%
  \institution{The Hong Kong Polytechnic University}
  \city{Hong Kong}
  \country{Hong Kong}}
\email{yanghong.zhou@connect.polyu.hk}
\orcid{0000-0002-0372-996X}

\author{Honghong He}
\affiliation{%
  \institution{The Hong Kong Polytechnic University}
  \city{Hong Kong}
  \country{Hong Kong}}
\email{21039747r@connect.polyu.hk}
\orcid{0000-0002-8766-7582}

 \author{P.~Y.~Mok}
\authornote{P.~Y.~Mok is the corresponding author (tracy.mok@polyu.edu.hk).}
\affiliation{%
  \institution{The Hong Kong Polytechnic University}
  \city{Hong Kong}
  \country{Hong Kong}}
\affiliation{
  \institution{Laboratory for Artificial Intelligence in Design}
  \city{Hong Kong}
  \country{Hong Kong}}
\email{tracy.mok@polyu.edu.hk}
\orcid{0000-0002-0635-5318}

\begin{abstract}
This paper reports on the development of \textbf{a novel style guided diffusion model (SGDiff)} which overcomes certain weaknesses inherent in existing models for image synthesis. The proposed SGDiff combines image modality with a pretrained text-to-image diffusion model to facilitate creative fashion image synthesis. It addresses the limitations of text-to-image diffusion models by incorporating supplementary style guidance, substantially reducing training costs, and overcoming the difficulties of controlling synthesized styles with text-only inputs. This paper also introduces a new dataset -- SG-Fashion, specifically designed for fashion image synthesis applications, offering high-resolution images and an extensive range of garment categories. By means of comprehensive ablation study, we examine the application of classifier-free guidance to a variety of conditions and validate the effectiveness of the proposed model for generating fashion images of the desired categories, product attributes, and styles. The contributions of this paper include a novel classifier-free guidance method for multi-modal feature fusion, a comprehensive dataset for fashion image synthesis application, a thorough investigation on conditioned text-to-image synthesis, and valuable insights for future research in the text-to-image synthesis domain. The code and dataset are available at: \url{https://github.com/taited/SGDiff}.
\end{abstract}



\begin{teaserfigure}
\includegraphics[width=0.88
\textwidth]{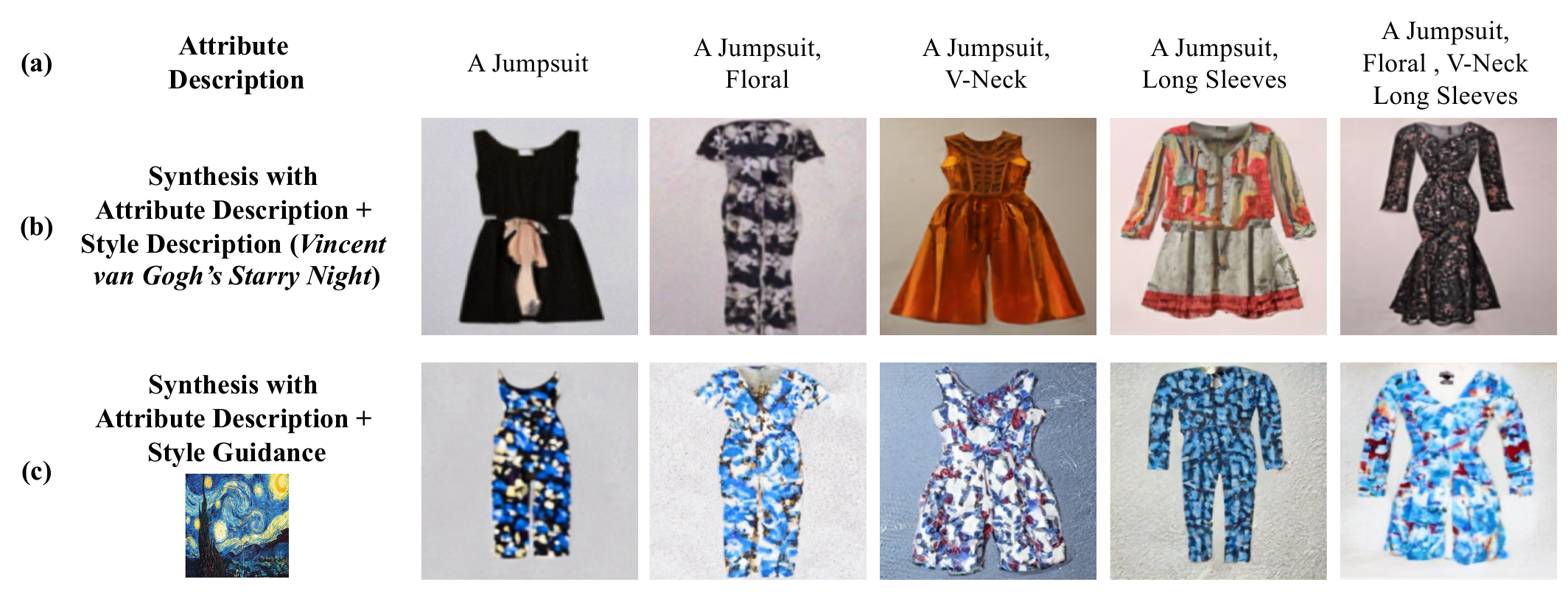}
\centering
  \vspace{-15px}
  \caption{Synthesizing clothing images with specific attributes and artistic styles: text-only synthesis results vs SGDiff results.}
  \vspace{3px}
  \label{fig:teaser}
\end{teaserfigure}


\maketitle

\section{Introduction}
Creative fashion synthesis, more specifically fashion image synthesis, holds enormous potential in the fashion industry, as it enables the generation of a diverse range of designs, which can inspire and foster innovation. By synthesizing fashion product images with specific attributes and styles, designers can rapidly explore a wide array of design concepts and ideas, reducing the time and resources needed for prototyping. For instance, Figure~\ref{fig:teaser} (row (a)) shows that a single jumpsuit category may encompass numerous attributes, including sleeve length, neckline type, prints, and overall styles. The creative fashion synthesis may involve an arbitrary combination of these attributes and styles, as shown in rows (b) and (c) of Figure~\ref{fig:teaser}. Even though the existing fashion synthesis approaches~\cite{ak2019attribute, hu2022spg, xu2021virtual, ge2021parser, cui2021dressing, lewis2021tryongan, kim2020style,jiang2022deep,yu2019personalized, zhou2022coutfitgan} can achieve promising results, they rely on different input modalities, such as category labels of textual modality, design sketches of visual modality, or other specialized inputs, which limit their ability to capture the large diversity of fashion product attributes and styles. 
Among the different input modalities, textual modality offers significant advantages in creative fashion synthesis since it provides a more accessible and flexible means to convey various garment categories and attribute information. Designers can easily communicate their design ideas by text descriptions, enabling a more streamlined design process and a broader exploration of garment variations. 

The recent advancement in Denoising Diffusion Probabilistic Models (DDPM)~\cite{ho2020denoising, nichol2021improved, dhariwal2021diffusion, song2021maximum, song2021denoising} has radically improved text-to-image generation, achieving stunning results with reasonable semantics. 
DDPMs can generate images that align with the input text descriptions, whereas the process of image synthesis is formulated as recovering the target image from an initial noisy image. 
Nichol~\emph{et al.}~\cite{nichol2022glide} proposed GLIDE, a UNet-like structure for posterior probability estimation in the denoising process, to incorporate text conditions so as to control synthesis directions. Furthermore, Rombach~\emph{et al.}~\cite{rombach2022high} investigated LDM model synthesizing high-resolution images with reasonable semantics using a Variational Autoencoder (VAE) to compress images into latent space and applying diffusion models to learn denoising in the latent space. Building on LDM~\cite{rombach2022high}, several studies~\cite{instructPix2Pix, hertz2023prompttoprompt} explored detailed image editing using text descriptions as control. Even though these approaches can effectively synthesize images with reasonable and desired semantics, there are considerable training costs involved. For example, the GLIDE was trained on several hundred million text-image pairs while the LDM was trained on the LAION-5B~\cite{schuhmann2022laionb} dataset. The huge model sizes limit the possible downstream applications. 
On the other hand, generating high-quality images that capture the essence of the desired design semantics, based only on text descriptions, is challenging due to the high dimensionality and variability in visual modality. The current methods could only control the synthesized results with simple descriptions, such as colors, but not visual styles like fabric textures, because describing different abstract styles with natural human languages is itself challenging.

In this paper, a novel approach called \textbf{S}tyle-\textbf{G}uide \textbf{Diff}usion Model (\textbf{SGDiff}) is developed to address the drawbacks of text-to-image diffusion models in fashion image synthesis. SGDiff is inspired by the old idiom \textit{"A picture is worth a thousand words"} and that style information could be better described and conveyed by images than by texts. Incorporating style guidance as control for image synthesis presents several challenges. Firstly, laborious data annotation is involved for selecting representative images as style guidance for the training purpose. Secondly, the alignment of features in different modalities remains a challenging problem when both style and text are simultaneously required to  control a model, especially for domain specific semantics not being well covered by large paired image-text dataset.
To circumvent the first challenge, we formulate the synthesis process as image reconstruction. By learning to reconstruct a garment from text descriptions and a randomly cropped image patch as style guidance, the proposed SGDiff can synthesize fashion garments that reflect both the text and style. 
To address the second issue, the proposed SGDiff model utilizes the image encoder of the Contrastive Language-Image Pre-Training (CLIP) model~\cite{CLIP} to convert style images into semantic representations. Furthermore, a Skip Cross-Attention (SCA) module is specially designed and applied to integrate image modality with text modality. 
Such network design is very different from the existing CLIP guided methods~\cite{patashnik2021styleclip,crowson2022vqgan,liu2021fusedream}, which align features in latent space using the distinct image and text encoders of CLIP and optimize the input latent variable as additional loss guidance. Existing methods suffer from the low extension ability for downstream applications and the optimization-based approach results in very slow image synthesis. 
Instead, with the help of the SCA module, we could fix the pretrained text-to-image diffusion and only fine tune the style (image) encoder and SCA module, significantly reducing the computational costs and addressing the multi-modal alignment problem.
Moreover, most existing diffusion models only consider applying classifier-free guidance based on a single condition, whereas SGDiff explores the optimized way of applying multi-condition classifier-free guidance to the diffusion model.
The key contributions of this paper are summarized as follows:
\begin{enumerate}[label=\roman*.,topsep=5pt,leftmargin=*]
\item A \textbf{new task} for creative fashion synthesis is addressed that both texts and style images are used to control the synthesis of fashion images under specific garment categories, attributes and styles. 
\item SGDiff -- a novel approach is developed that integrates image modality to a pretrained text-to-image diffusion model, enabling creative fashion synthesis with style guidance. To the best of our knowledge, this is \textbf{the first network proposal} integrating CLIP with the classifier-free guidance approach for modality fusion aiming toward conditioned image generation. 
\item With the innovative network design, \textbf{a new network training strategy} is presented that significantly reduces training costs, only requiring fine-tuning the image encoder and modality fusion module rather than the entire network.
\item A \textbf{SG-Fashion dataset} is specifically constructed, which features high-resolution images and covers a wide range of garment categories. The proposed method has been validated both on this SG-Fashion and the Polyvore datasets.
\item This is \textbf{the first of its kind of thorough investigation for extending classifier-free guidance to multiple conditions}, providing valuable insights for future research in the text-to-image synthesis domain.
\end{enumerate}

\section{Related Work}
\subsection{Fashion Synthesis}
Fashion synthesis, an emerging research area within the broader field of computer vision and generative models, concentrates on generating and manipulating fashion-related images, such as clothing and accessories as well as fashion models. 
Virtual try-on (VTON) has generated considerable attention in some recent studies~\cite{hu2022spg, xu2021virtual, ge2021parser, cui2021dressing, lewis2021tryongan, kim2020style}, which typically employ human parsing maps and pose estimation techniques to transfer textures from a desired garment onto a target person. Although these VTON approaches successfully synthesize consistent clothing attributes, they primarily focus on human-centric scenarios.

Several recent studies have investigated garment-centric fashion synthesis, with the aim to generate novel and diverse clothing items. For example, Jiang \emph{et al.}~\cite{jiang2022deep} developed FashionG to transfer styles onto a garment without changing its original image content.
Other researchers~\cite{yu2019personalized,zhou2022coutfitgan,ding2023personalized} explored the synthesis of compatible fashion based on a given garment image as a query.
\textit{These aforementioned studies are all using visual modality input as control for image synthesis, their ability to control the detail attributes of the generated fashion is rather limited.}

Text-to-image fashion synthesis remains relatively unexplored compared to other fashion synthesis approaches. Zhu~\emph{et al.}~\cite{zhu2017your} proposed a method that uses textual descriptions to edit images of garments worn by humans. Zhang~\emph{et al.}~\cite{zhang2022armani}developed an ARMANI model for fashion synthesis based on multi-modal inputs including text descriptions and edge or regional detail in image modality. \textit{Although the above approaches successfully enable control over the synthesized garments, they generally fail to achieve detailed control of the synthesized textures or styles.}

\subsection{CLIP Model Guided Modality Fusion}
The CLIP model, introduced by OpenAI \cite{CLIP}, has revolutionized the field of computer vision by leveraging the power of large-scale transformers trained on both images and text. One of the main strengths of the CLIP model is its zero-shot learning capability, namely no learning is needed, which allows it to handle new tasks without requiring any task-specific fine-tuning. Its zero-shot capability has been exploited in various applications, such as image classification \cite{zhang2022tip, esmaeilpour2022zero}, object detection \cite{teng2021global, shi2022proposalclip}, and semantic segmentation \cite{liang2022open, zhou2022zegclip, zhou2022extract}.

CLIP models have been integrated with generative models like GANs \cite{liu2021fusedream, StyleCLIPDraw} and VQ-VAEs \cite{crowson2022vqgan} to produce impressive results in various tasks, from text-to-image synthesis to image editing. 
For example, StyleCLIP \cite{patashnik2021styleclip} utilizes a pretrained StyleGAN \cite{karras2020analyzing} and the CLIP model to align image and text features within the style space. VQGAN-CLIP \cite{crowson2022vqgan} uses CLIP as additional guidance to control the generation direction in pretrained generative model. FuseDream \cite{liu2021fusedream} is a training-free method integrating the latent generation space with CLIP embeddings.
DALL$\cdot$E \cite{ramesh2021zero} combines the CLIP model with a discrete VAE to generate high-quality images from textual descriptions. \textit{All these models adopt a training-free pipeline and treat the CLIP model as a gradient guidance to interpret the generation of latent space. Although these methods could integrate pretrained generation models with CLIP for text-to-image synthesis, they synthesize every image as a separate optimization process, which are computationally costly, and they fail to capture domain-specific text descriptions.}

\subsection{Text-to-Image Diffusion Models}
Diffusion models have recently emerged as a powerful branch of generative models, demonstrating their superior capabilities of handling image, text, audio as well as other modalities of data~\cite{meng2022sdedit, su2023dual, nichol2022glide, rombach2022high, leng2022binauralgrad}. These models aim to learn the data distribution by performing a Markov chain, simulating the data generation process in reverse~\cite{ho2020denoising, nichol2021improved, dhariwal2021diffusion, song2021maximum, song2021denoising}. 

Despite the many research studies are focusing on synthesizing high-resolution images using diffusion models, there is a growing body of research that is interested in more controlled synthesis.  Hertz~\emph{et al.}~\cite{hertz2023prompttoprompt} investigated a Prompt-to-Prompt mechanism of text-to-image generation, where text features activate feature maps through cross-modal attention. 
InstructPix2Pix \cite{instructPix2Pix} combines the large pretrained language model GPT3 \cite{GPT3} and the state-of-the-art text-to-image LDM~\cite{rombach2022high} model to synthesize a dataset for text-driven image editing. 
\textit{Although these methods can synthesize images with corresponding semantics, they are trained on large open-domain datasets and have difficulty in capturing terms specific to the fashion domain.}
Recently, Textual Inversion \cite{gal2023an} and DreamBooth \cite{ruiz2023dreambooth} can adapt pre-trained diffusion models with new styles. \textit{Model retraining is, however, needed for every new style.}  

\begin{figure*}[!t]
    \centering
    \includegraphics[width=0.92\textwidth]{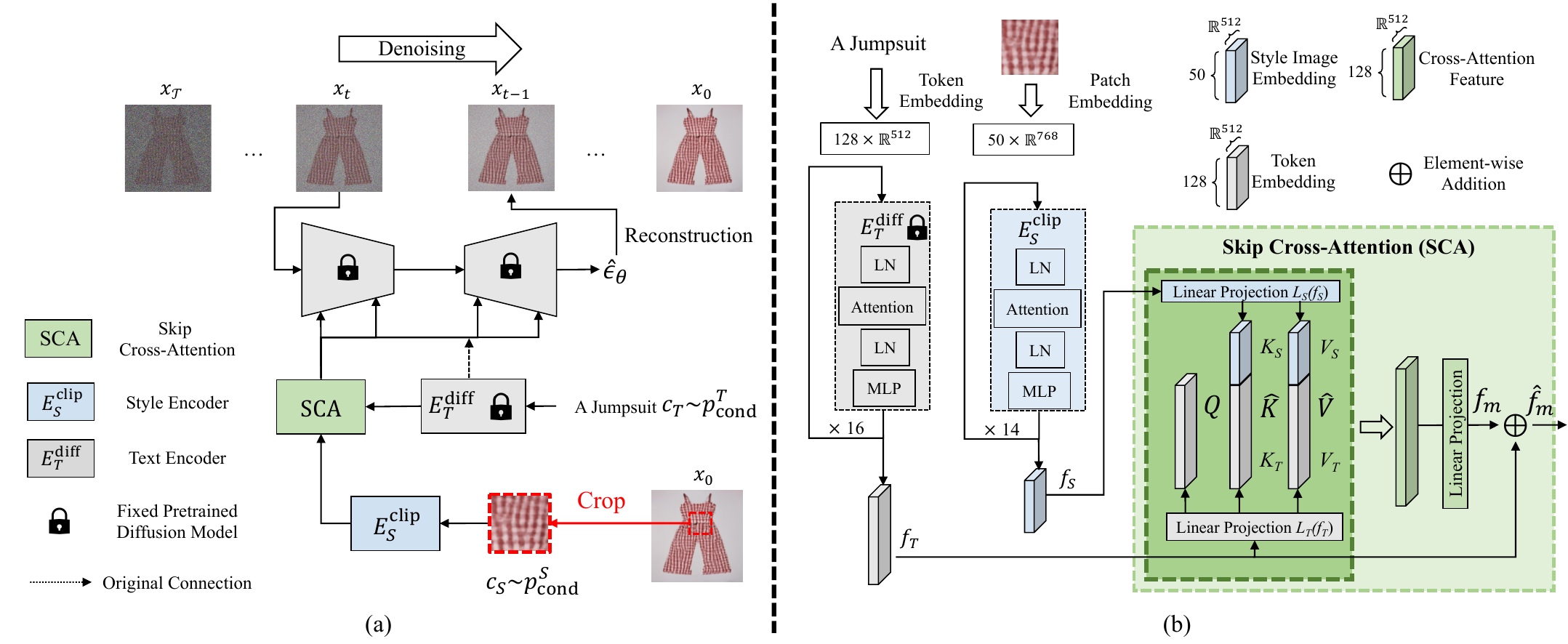}
    \vspace{-10px}
    \caption{The proposed Style-Guided Diffusion Model (SGDiff) network (a) an overview and (b) detail model: SGDiff takes two inputs, a text condition ($c_T$) for garment attributes and a style condition ($c_S$) for style guidance, and leverages the Skip Cross-Attention (SCA) module and a pretrained CLIP image encoder for efficient training and resource utilization.}
    \label{fig:model}
\end{figure*}
\section{Method}
\subsection{Preliminaries}
\label{sec:preliminaries}
Diffusion models utilize a Markov chain process, motivated by non-equilibrium thermodynamics, to simulate forward diffusion process.
Given an image $\mathbf{x}_0 \sim q(\mathbf{x})$, the forward diffusion process adds small amount of Gaussian noises to $\mathbf{x}_0$ in $\mathcal{T}$ steps, producing a set of noisy images $\mathbf{x}_1,...,\mathbf{x}_\mathcal{T}$. The step sizes $\beta_t$ are controlled by a variance schedule $\left\{\beta_t \in(0,1)\right\}_{t=1}^\mathcal{T}$:
\begin{subequations}
\begin{equation}
    q\left(\mathbf{x}_{1: \mathcal{T}} \mid \mathbf{x}_0\right)=\prod_{t=1}^\mathcal{T} q\left(\mathbf{x}_t \mid \mathbf{x}_{t-1}\right)
\end{equation}
\begin{equation}
q\left(\mathbf{x}_t \mid \mathbf{x}_{t-1}\right)=\mathcal{N}\left(\mathbf{x}_t ; \sqrt{1-\beta_t} \mathbf{x}_{t-1}, \beta_t \mathbf{I}\right).
\end{equation}
\end{subequations}
By reparameterization, let $\alpha_t=1-\beta_t$ and $\bar{\alpha}_t=\prod_{i=1}^t \alpha_i$, the forward process can sample $\mathbf{x}_t$ at arbitrary timestep $t$ directly from $\mathbf{x}_0$:
\begin{equation}
    q\left(\mathbf{x}_t \mid \mathbf{x}_0\right)=\mathcal{N}\left(\mathbf{x}_t ; \sqrt{\bar{\alpha}_t} \mathbf{x}_0,\left(1-\bar{\alpha}_t\right) \mathbf{I}\right).
\end{equation}
When $\mathcal{T} \rightarrow \infty$, the image $\mathbf{x}_0$ will be diffused to a standard Gaussian noise $\mathbf{x}_\mathcal{T} \sim \mathcal{N}(0,\mathbf{I})$. 
Given a Gaussian noise, a neural network model is then learned to approximate the conditional probabilities to reverse the diffusion process $p_\theta$ as follows:
\begin{subequations}
\begin{equation}
p_\theta\left(\mathbf{x}_{0: \mathcal{T}}\right)=p\left(\mathbf{x}_\mathcal{T}\right) \prod_{t=1}^\mathcal{T} p_\theta\left(\mathbf{x}_{t-1} \mid \mathbf{x}_t\right)
\end{equation}
\begin{equation}
p_\theta\left(\mathbf{x}_{t-1} \mid \mathbf{x}_t\right)=\mathcal{N}\left(\mathbf{x}_{t-1} ; \boldsymbol{\mu}_\theta\left(\mathbf{x}_t, t\right), \boldsymbol{\Sigma}_\theta\left(\mathbf{x}_t, t\right)\right),
\label{eq:reverse_p}
\end{equation}
\end{subequations}
where $\boldsymbol{\mu}_\theta$ and $\boldsymbol{\Sigma}_\theta$ are the approximated mean and variance of the reversed Gaussian distribution.
By simplifying $\boldsymbol{\Sigma}_\theta$ as constant $\beta_t$, $\boldsymbol{\mu}_\theta$ is tractable~\cite{ho2020denoising}:
\begin{equation}
\boldsymbol{\mu}_\theta\left(\mathbf{x}_t, t\right) =\frac{1}{\sqrt{\alpha_t}}\left(\mathbf{x}_t-\frac{1-\alpha_t}{\sqrt{1-\bar{\alpha}_t}} \boldsymbol{\epsilon}_t\right).
\label{eq:mu}
\end{equation}

With $\mathbf{x}_t$ known during training, the network $\boldsymbol{\epsilon}_\theta$ is reparameterized to predict noise $\epsilon_t$ from input $\mathbf{x}_t$ at time step $t$ with this simplified objective~\cite{ho2021classifierfree}:

\begin{equation}
\label{eq:diffusion_loss}
\mathcal{L}_t^{\text{simple}} =\mathbb{E}_{t \sim[1, \mathcal{T}], \mathbf{x}_0, \epsilon_t}\left[\left\|\epsilon_t-\boldsymbol{\epsilon}_\theta\left(\sqrt{\bar{\alpha}_t} \mathbf{x}_0+\sqrt{1-\bar{\alpha}_t} \epsilon_t, t\right)\right\|^2\right].
\end{equation}
For brevity, $\boldsymbol{\epsilon}_\theta(x_t, t)$ is denoted as $\boldsymbol{\epsilon}_\theta(x_t)$ hereafter in this article.

\subsection{SGDiff Overview}
SGDiff aims to achieve detailed control over synthesized fashion images in terms of both correct garment attributes and garment textures (styles). 
Controlling detailed garment textures using natural language is challenging, therefore, the proposed SGDiff, as illustrated in Figure~\ref{fig:model}, takes two inputs: a text condition ($c_T$) describing the garment attributes and a style condition ($c_S$) guiding the synthesized garment texture.
The text encoder $E_T^{\text{diff}}$ of the diffusion model encodes the semantic representation $f_T$, and the style encoder $E_S^{\text{clip}}$ of a pretrained CLIP model encodes the style representation $f_S$. The diffusion network $\epsilon_\theta$ estimates the noise $\hat{\epsilon}_t$ as follows: 
\begin{equation}
\hat{\epsilon}_t = \epsilon_\theta \left( x_t, E_T^{\text{diff}}(c_T), E_S^{\text{clip}}(c_S) \right).
\label{eq:overview}
\end{equation}

To avoid labor-intensive data annotation, the conditioned image synthesis is formulated as an image reconstruction task, as shown in Figure~\ref{fig:model}(a), in which a randomly image patch cropped from the garment image is taken as style condition $c_S$, the model is then trained to reconstruct garment according to the style guidance $c_S$.

To achieve efficient training, we have the pre-trained text-to-image diffusion model fine-tuned on a domain-specific dataset using text as input condition, according to a classifier-free guidance approach~\cite{ho2021classifierfree}. Next, by fixing the diffusion network parameters, we optimize the specially designed SCA module and fine-tune a pretrained image encoder $E_S^{\text{clip}}$ with multiple conditions of text description and style guidance, which will be discussed in detail in Section~\ref{sec:classifier-free}. 

\begin{figure*}[htbp]
    \centering
    \includegraphics[width=0.92\textwidth]{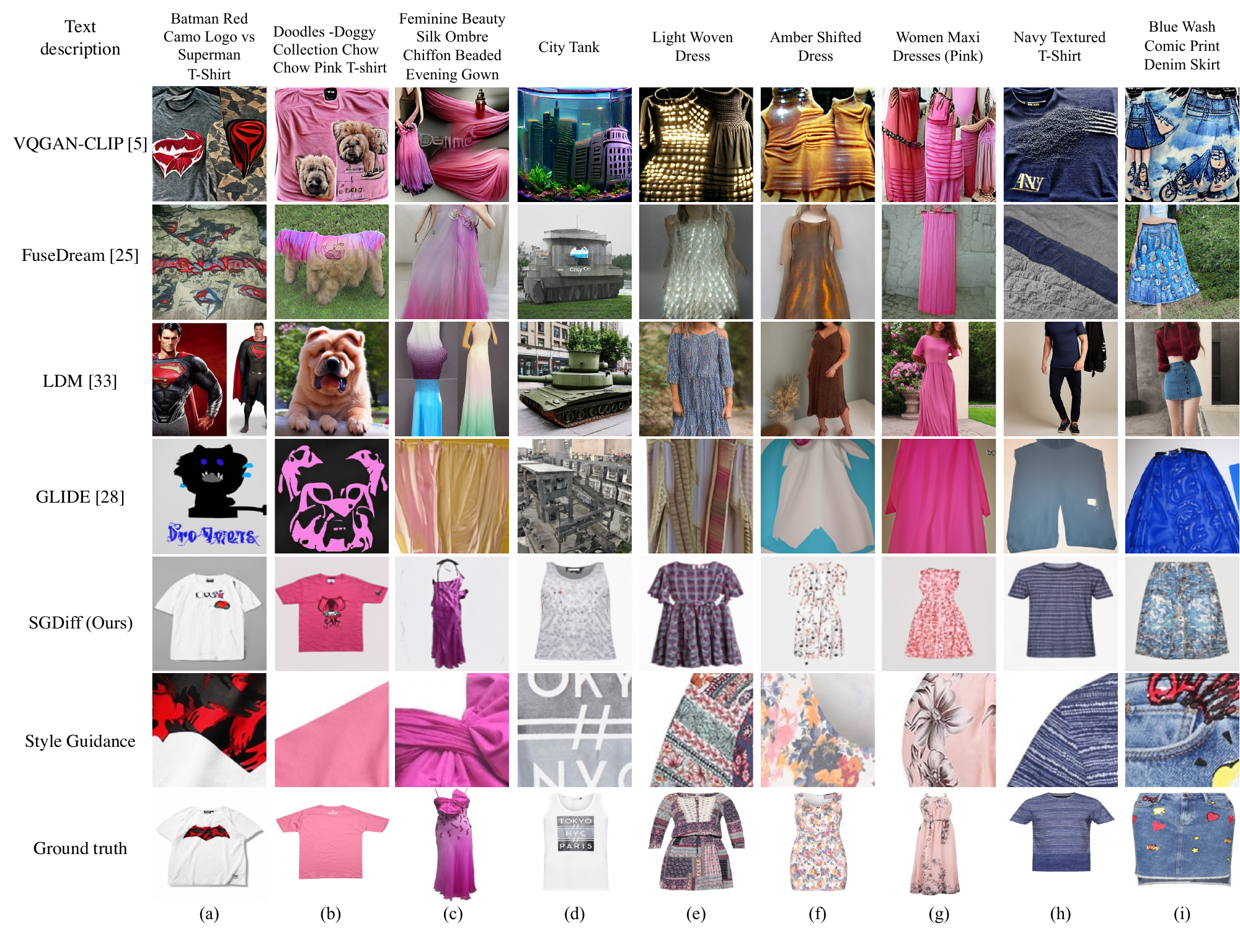}
    \vspace{-15px}
    \caption{Qualitative comparison of SGDiff with state-of-the-art (SOTA) approaches. The 2nd and 3rd rows illustrate the results of CLIP-based methods of VQGAN-CLIP \cite{crowson2022vqgan} and FuseDream \cite{liu2021fusedream}, while the 4th and 5th rows illustrate the results of diffusion-based methods of LDM \cite{rombach2022high} and GLIDE \cite{nichol2022glide}. The 6th row illustrates SGDiff's ability to incorporate style images (the 7th row) into text conditions (the 1st row), successfully synthesizing garments with the desired textures.}
    \label{fig:qualitative}
\end{figure*}

\subsection{Skip Cross-Attention Module}
Figure~\ref{fig:model}(b) illustrates the process of integrating two different modalities, namely text description of garment attributes $c_T$ and image of style guidance $c_S$, in the proposed SGDiff model. The integration of the two input modalities is achieved through the specially designed Skip Cross-Attention (SCA) module.

Both encoders, $E_T^{\text{diff}}$ and $E_S^{\text{clip}}$, employ transformer-based structures and the output features $f_T \in\mathbb{R}^{128 \times 512}$ and $f_S \in \mathbb{R}^{50 \times 512}$ represent two modalities of input. Such aligned features of $f_T$ and $f_S$ enable easy integration of the two representations by attention mechanism \cite{ashish2017attention}. 
To do so, the semantic representation $f_T$ is linearly projected into query and key-value pairs:
\begin{equation}
    Q, K_T, V_T = L_T(f_T),
\end{equation}
where $L_T$ represents linear projection, and query $Q$ and key-value pairs ${K_T, V_T}$ all have size $\mathbb{R}^{128 \times 512}$. The style representation $f_S$ is projected into key-value pairs only:
\begin{equation}
    K_S, V_S = L_S(f_S).
\end{equation}
The style key-value pairs are concatenated with text key-value pairs:
\begin{equation}
\hat{K} = K_S~(+)~K_T \qquad \text{and}  \qquad   \hat{V} = V_S~(+)~V_T,
\end{equation}
where $(+)$ denotes length-wise concatenation. 

Specifically, the semantic representation $f_T$ is chosen as query $Q$ because it provides key attribute information for garment synthesis. With $f_T$ as query, style representation $f_S$ is aligned with the garment attributes in order to improve the quality of the synthesized images.
The cross-attention is implemented by integrating the key-value pairs from both modalities as follows:
\begin{equation}
    f_m = \operatorname{Attention}(Q, \hat{K}, \hat{V})=\operatorname{softmax}\left(\frac{Q \hat{K}^T}{\sqrt{d_k}}\right) \hat{V}.
    \label{eq:CA}
\end{equation}
Finally, the skip connection is applied, as shown in Figure~\ref{fig:model}:
\begin{equation}
    \hat{f_m} = f_m + f_T.
    \label{eq:skip}
\end{equation}
The SCA module enables effective integration of text and image modalities, allowing the SGDiff model to control the synthesized texture without any reduction in semantic control.

\subsection{Training Objectives}
As discussed in Section~\ref{sec:preliminaries}, diffusion models implicitly learn to reconstruct an image from Gaussian noise. The network $\epsilon_\theta$ estimates the noise in the current input noisy image $\mathbf{x}_t$. The training objective of DDPM (Eq. (\ref{eq:diffusion_loss})), however, does not address condition constraints explicitly. 
Therefore, SGDiff employs perceptual loss, in addition to Eq. (\ref{eq:diffusion_loss}), to govern image synthesis. To this end, the reconstructed image $\hat{x}_0$ is obtained at every time step $t$, according to the estimated noise $\hat{\epsilon_t}$ by Eq. (\ref{eq:overview}):
\begin{equation}
\hat{\mathbf{x}_0}=\frac{1}{\sqrt{\bar{\alpha}_t}}\left(x_t-\sqrt{1-\bar{\alpha}_t} \hat{\epsilon_t}\right).
\end{equation}
The Perceptual Loss~\cite{johnson2016perceptual} is then calculated by:
\begin{equation}
\label{eq:perc_loss}
\mathcal{L}_t^{\text {perc }}=\mathbb{E}_{m}\left\|\boldsymbol{\psi}_m\left(\hat{\mathbf{x}}_0\right)-\boldsymbol{\psi}_m\left(\mathbf{x}_0\right)\right\|_2,
\end{equation}
where $\boldsymbol{\psi}_m$ denotes the $m$-th layer of VGG. Following~\cite{johnson2016perceptual}, the layers of \textit{relu1\_2, relu2\_2, relu3\_2, relu4\_2,} and \textit{relu5\_2} are used in Eq. (\ref{eq:perc_loss}).
The overall training objective with Perceptual Loss, adapted from \cite{nichol2021improved}, is as follows:
\begin{equation}
\label{eq:overall_loss}
    \mathcal{L} = \lambda_s \mathcal{L}_t^{\text {simple}} + \mathcal{L}_t^{\text {vlb}} + \lambda_p \mathcal{L}_t^{\text{perc}},
\end{equation}
where $\lambda_s$ and $\lambda_p$ are balancing weights for the corresponding losses.

\subsection{Multi-Modal Conditions}
\label{sec:classifier-free}
Classifier-free guidance~\cite{ho2021classifierfree} has obvious advantages over classifier guidance~\cite{dhariwal2021diffusion} for conditioned generation with DDPMs. For more flexible control, the proposed SGDiff also adopts classifier-free guidance approach~\cite{ho2021classifierfree}, in which the model $\epsilon_\theta$ is trained with conditional state $c$ and unconditional state $\varnothing$ according to a certain probability $c \sim p_\text{cond}$:
\begin{equation}
\hat{\epsilon_\theta}\left(x_t, c\right)=
\epsilon_\theta\left(x_t, \varnothing\right)
+ s \left[\epsilon_\theta\left(x_t, c\right)-\epsilon_\theta\left(x_t, \varnothing\right) \right].
\label{eq:classifier-free}
\end{equation}

Nevertheless, the above approach (Eq. (\ref{eq:classifier-free})) does not address more complex situation where conditions are multiple, happen in different combinations at varied probabilities. Until recently, InstrucPix2Pix~\cite{instructPix2Pix} suggested different weights for two conditions:
\begin{equation}
\label{eq:multi-classifier-free}
\begin{aligned}
\hat{\epsilon_\theta}\left(x_t, c_1, c_2\right)= & \epsilon_\theta\left(x_t, \varnothing, \varnothing\right) \\
& +s_1 \left[\epsilon_\theta\left(x_t, c_1, \varnothing\right)-\epsilon_\theta\left(x_t, \varnothing, \varnothing\right)\right] \\
& +s_2 \left[\epsilon_\theta\left(x_t, c_1, c_2\right)-\epsilon_\theta\left(x_t, c_1, \varnothing\right)\right],
\end{aligned}
\end{equation}
where $s_1$ and $s_2$ indicate the weight scale of condition $c_1 \sim p^1_{\text{cond}}$ and $c_2 \sim p^2_{\text{cond}}$, respectively. 
In~\cite{instructPix2Pix}, however, it was not discussed either the order of $c_1$ and $c_2$ or the weight scales $s_1$ and $s_2$. 

In the current task, Eq.~(\ref{eq:multi-classifier-free}) is applied by setting the two conditions as $c_T$ and $c_S$. The SGDiff is subjected to two conditions with independent conditional probability $p^S_{\text{cond}}=0.8$ and  $p^T_{\text{cond}}=0.8$. In model training, like all text-to-image diffusion models, the unconditional state $\varnothing$ of textual condition $c_T$ is set to padding token.
The unconditional state $\varnothing$ of style guidance $c_S$ is done by inputting a blank (background only) patch image. 

\noindent\textbf{Background masking:} Apart from inputting a blank image patch as unconditional state, the background color in RGB space may also appear in the foreground. To avoid confusion, we mask the background pixel values to -255 to distinguish them from the normal RGB values. Such masking technique allows the model to focus more on the foreground texture. The effectiveness of such background masking setting will be evaluated in Section~\ref{sec:ablation}. 

\noindent\textbf{Condition order and weight scales:} In order to explore the effect of the condition order, by setting $c_1=c_S$ and $c_2=c_T$, alternatively $c_1=c_T$ and $c_2=c_S$, in Eq.~(\ref{eq:multi-classifier-free}), and $s_T=1$, this will result in
\begin{subequations}
\begin{equation}
\begin{split}
    \hat{\epsilon}_\theta(x_t, c_S, c_T) & = (s_S-1)\left[\epsilon_\theta(x_t, c_S, \varnothing) - \epsilon_\theta(x_t, \varnothing, \varnothing \right] \\
    &  + \epsilon_\theta(x_t, c_S, c_T) 
    \label{eq:image prior to text}
\end{split}
\end{equation}
\begin{equation}
\begin{split}    
\hat{\epsilon}_\theta(x_t, c_T, c_S) & =  (s_S-1) \left[\epsilon_\theta(x_t, c_T, c_S) - \epsilon_\theta(x_t, c_T, \varnothing) \right] \\
    & + \epsilon_\theta(x_t, c_T, c_S).
    \label{eq:text prior to image}
\end{split}
\end{equation}    
\end{subequations}

In our implementation, the model $\epsilon_\theta$ takes $c_S$ and $c_T$ simultaneously, the two terms $\epsilon_\theta\left(x_t,c_S,c_T\right)$ and $\epsilon_\theta\left(x_t,c_T,c_S\right)$ are therefore equivalent.
Comparing Eq. (\ref{eq:image prior to text}) wtih (\ref{eq:text prior to image}), thus $[\epsilon_\theta(x_t, c_S, \varnothing) - \epsilon_\theta(x_t, \varnothing, \varnothing)]$ = $[\epsilon_\theta(x_t, c_T, c_S) - \epsilon_\theta(x_t, c_T, \varnothing)]$.
It implies that if the style condition and text condition are independent, the condition order will not have a significant impact on the image generation. 
Moreover, the weight scale serves to adjust the influence of style guidance. When $s_S>s_T$ (i.e. $s_S>1$ when $s_T=1$), it introduces a positive conditioned direction to the denoising processing, emphasizing the influence of condition is guiding the synthesis. The multi-condition synthesis will be further evaluated in Section~\ref{sec:ablation}.

\section{Experiments}
\subsection{Datasets and Implementation Details}
In this study, we prepared a SG-Fashion dataset with 17,000 fashion product images downloaded from e-commerce websites including ASOS, Uniqlo and H\&M. We set aside 1,700 images as the test set. The dataset covers 72 product categories, encompassing most types of garment items. Since our SGDiff does not rely on textural descriptions, we use the original product titles as text descriptions. Apart from the SG-Fashion dataset, we also experimented on the publicly available dataset of Polyvore~\cite{han2017learning} with the same settings.

GLIDE~\cite{nichol2022glide} was adopted as the backbone text-to-image diffusion model, which uses a low-resolution generation model for size $64\times 64$ and a super-resolution model to upsample the generated low-resolution image to the size of $256 \times 256$. We fine tuned the generation model and directly employed the super-resolution model as the pretrained text-to-image model. For the pretrained CLIP image encoder, we chose vision transformer of ViT/32. To speed up the synthesis process, we adopted DDIM \cite{song2021denoising} scheduler with 100 sapmling steps for all diffusion-based models.

The backbone model (GLIDE) was fine tuned on the domain-specific dataset that
the AdamW optimizer was used with a learning rate of $1e^{-4}$, and the model was optimized for 235,000 iterations. Due to GPU limitations, we set the batch size to 8 and trained the GLIDE on a single RTX 3090 GPU. We also used AdamW but with a learning rate of $1e^{-5}$ for training the SGDiff with 50,000 iterations for all experiments on a single RTX 3090 GPU. In terms of the SCA module, we adopted multi-head attention with $4$ heads. In all experiments, we set $\lambda_s = 1$ and $\lambda_p = 0.001$ in Eq. (\ref{eq:overall_loss}). 
Since the training of SGDiff fixes the parameters of the pretrained backbone, we can use a larger batch size of 16. 
For SGDiff training, we cropped a single texture patch from the foreground. To ensure this cropped patch provides sufficient style information, we applied BASNet~\cite{qin2019basnet} for background masking.

\begin{figure}[t]
    \centering
    \includegraphics[width=0.45\textwidth]{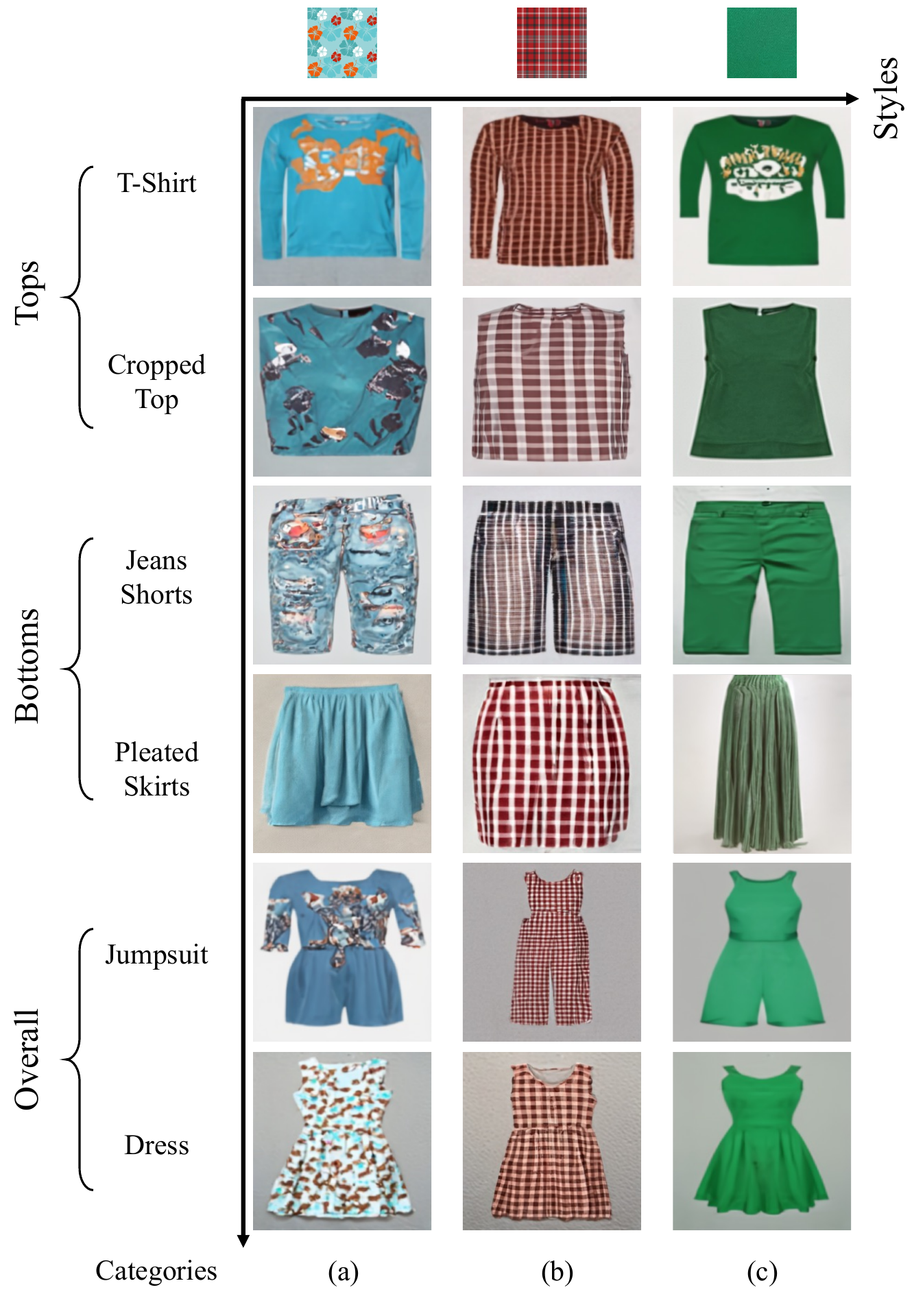}
    \vspace{-10px}
    \caption{Illustration of SGDiff's capability to synthesize garments across various categories and styles, using style guidance of different colors.}
    \label{fig:various garments}
    \vspace{-10px}
\end{figure}

\subsection{Qualitative Evaluation}
The qualitative evaluation compares the SGDiff results with several SOTA text-to-image generation methods, including LDM~\cite{rombach2022high} and GLIDE~\cite{nichol2022glide} for diffusion-based methods, and FuseDream \cite{liu2021fusedream} and VQGAN-CLIP \cite{crowson2022vqgan} for CLIP-guided GAN-based methods. All selected SOTA methods have zero-shot capability.  Figure~\ref{fig:qualitative} presents a comprehensive qualitative comparison of these methods. Generally speaking, FuseDream and LDM could synthesize garments in most cases, while VQGAN-CLIP and GLIDE could only synthesize fabrics. The proposed SGDiff could successfully implement the fashion synthesis with desired clothing category and style. Specifically, when synthesizing a garment with complex text descriptions (see examples in columns (a), (b), and (c)), the other methods tend to ignore the key message but capture part of the semantics like Batman logo, pink doggy, or silk, while SGDiff tends to synthesize clothing and consider the style guidance to control the synthesized textures.
Moreover, semantic confusion is one of main challenges in text-to-image synthesis. For instance, `Tank' refers to a specific type of upper clothing in the fashion domain. Column (d) of Figure~\ref{fig:qualitative} shows that both the diffusion-based and CLIP-based approaches have difficulty in capturing domain-specific semantics, 
while only SGDiff could synthesize a tank garment with specified styles.
The other columns present cases when offering textual descriptions like amber, light and pink, although the other SOTA methods could synthesize clothing with textures that are similar to the descriptions, they show greater differences to the ground truth images comparing to SGDiff. In conclusion, SGDfiff is suitable for fashion synthesis since it could capture the garment category and desired styles. 

In addition to the comparative analysis, Figure~\ref{fig:various garments} illustrates the innovative capability of SGDiff in synthesizing garments across various categories and styles. With style guidance images under different color schemes, SGDiff effectively transfers styles from the guidance images to the synthesized garments, meeting the condition of garment attributes. Figure~\ref{fig:various garments} shows a range of synthesized fashion under specific color scheme in each column, offering valuable inspiration for innovative fashion design. When conditioned generation are out of the training set, SGDiff can still exhibit a remarkable generative capability by successfully blending different condition combinations, e.g., the jeans shorts with red check and green patterns showed in columns (b) and (c) are not existed in the training data. Moreover, the style guidance appears in interesting variations in the generated fashion. These results highlight the versatility and robustness of the SGDiff model in the realm of fashion design.

\begin{table}[t]
\begin{threeparttable}
    \centering
    \caption{Quantitative evaluation and comparison of various SOTA methods} 
    \vspace{-10px}
    \begin{tabular}{c@{\hspace{0.5em}}c@{\hspace{0.5em}}c@{\hspace{0.5em}}c@{\hspace{0.5em}}c@{\hspace{0.5em}}c@{\hspace{0.5em}}c@{\hspace{0.5em}}}
    \toprule
    Datasets & \multicolumn{3}{c}{SG-Fashion} & \multicolumn{3}{c}{Polyvore} \\
    \cmidrule(lr){1-1} \cmidrule(lr){2-4} \cmidrule(lr){5-7}
    Metrics & LPIPS ↓ & FID ↓ & CS ↑ & LPIPS ↓ & FID ↓ & CS ↑ \\
    \midrule  
    VQGAN-CLIP \cite{crowson2022vqgan}  & 0.7364 & 95.84 & 22.20  & 0.7122 & 68.01 & \textbf{39.65} \\
    FuseDream \cite{liu2021fusedream} & 0.7067 & 60.44 & \textbf{38.03} & 0.7032 & \textbf{41.94} & 38.53$^*$ \\
    \hdashline
    LDM \cite{rombach2022high} & 0.7158 & 85.73 & 31.66$^*$ & 0.7214 & 59.79 & 31.89 \\
    GLIDE \cite{nichol2022glide} & 0.6921  & 78.7 & 23.72 & 0.7164 & 63.85 & 23.28\\
    \hdashline
    Ground Truth & -- & -- & \textit{29.13} & -- & -- & \textit{29.88}\\    
    Baseline & 0.5772$^*$ & 36.13$^*$ & 27.31 & 0.6637$^*$ & 43.5 & 26.24\\
    SGDiff (Ours)  & \textbf{0.4474} & \textbf{32.06} & 27.53 &
    \textbf{0.6369} & 41.98$^*$ & 27.33 \\
    \bottomrule
    \end{tabular}%
  \label{tab: quantitative}%
      \begin{tablenotes}
      \setlength\labelsep{0pt}
    \footnotesize
        \item the best results are in \textbf{bold}, and the second best results are indicated with $^*$.
   \end{tablenotes}
  \vspace{-10px}
\end{threeparttable}
\end{table}
\begin{table}[htbp]
  \centering
  \caption{Consumption of synthesizing an image with resolution of $256\times256$ on a RTX 3090 GPU}
  \vspace{-5px}
    \begin{tabular}{c@{\hspace{0.5em}}c@{\hspace{0.5em}}c@{\hspace{0.5em}}c@{\hspace{0.5em}}c@{\hspace{0.5em}}c@{\hspace{0.5em}}}
    \toprule
     & VQGAN-CLIP & FuseDream & LDM   & GLIDE & Ours \\
    \midrule
    Time  & 62s   & 171s  & 5.9s  & 9s    & 9.8s \\
    Memory & 5686M & 9296M & 6570M & 5550M & 5986M \\
    \bottomrule
    \end{tabular}%
  \label{tab:time}%
\end{table}%

\begin{figure*}
    \includegraphics[width=0.95\textwidth]{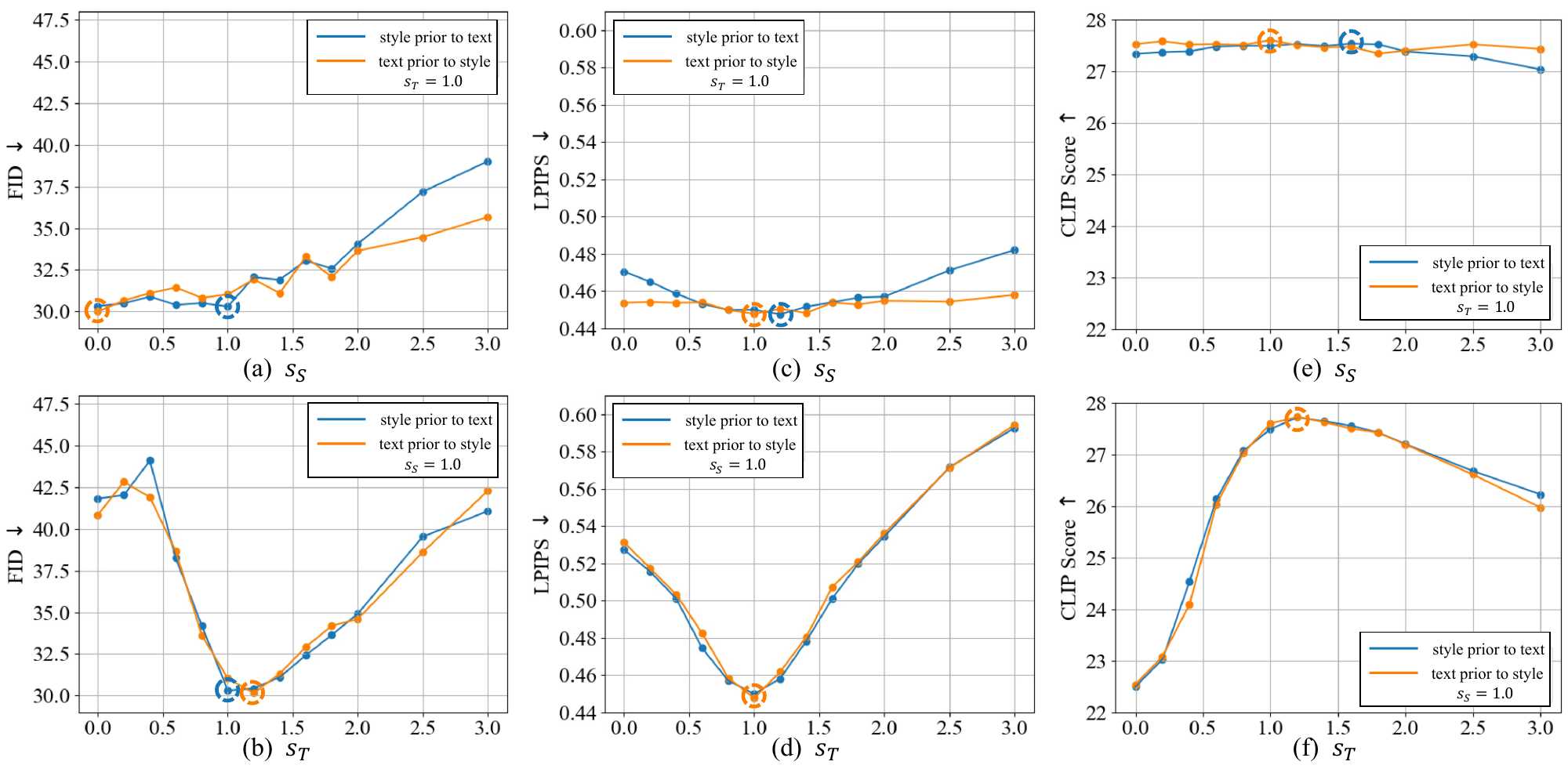}
    \vspace{-10px}
      \caption{Ablation study on the impact of style and text guidance on the performance of SGDiff in terms of (a) and (b) for FID, (c) and (d) for LPIPS and (e) and (f) for CLIP-score. We set one conditional weight varies in range of $[0, 0.2, 0.4, 0.6, 0.8, 1.0, 1.2, 1.4, 1.6, 1.8, 2.0, 2.5, 3.0]$ while the other conditional weight is fixed at $1.0$.}
  \label{fig:ablation}
\end{figure*}
\subsection{Metrics and Quantitative Evaluation}
Table~\ref{tab: quantitative} shows the quantitative evaluation, in which three metrics, including FID~\cite{HeuselFID}, LPIPS~\cite{Zhang_2018_CVPR} and CLIP-Score (CS)~\cite{CLIP}, are used to assess and compare the performance of SGDiff with other SOTA methods. FID and LPIPS measure the distance in feature space, with FID focusing on the overall distribution statistics of the generated/synthesized images and the ground truths, while LPIPS computes the distance between each pair of synthesized image and the corresponding ground truth, lower the FID and LPIPS values higher the image quality. In contrast, the CLIP-score measures the semantic correspondence, namely the cosine similarity between synthesized images and their corresponding text descriptions, with higher scores indicating better alignment.

As shown in Table~\ref{tab: quantitative}, SGDiff model performs the best in terms of LPIPS, comparing to other SOTA methods on both SG-Fashion and Polyvore datasets. SGDiff's FID value is also the lowest for SG-Fashion dataset and only slightly lower than FuseDream for Polyvore dataset by 0.04\%. This demonstrates that the SGDiff model can generate better images fulfilling the conditions without sacrificing the image quality. The CS of the SGDiff is higher than GLIDE and the baseline (i.e. GLIDE being fine tuned on the datasets), but lower than FuseDream and LDM, because FuseDream optimizes the BigGAN-256 \cite{brock2018large} latent space using CLIP guidance and LDM leverages a vast text-to-image dataset consisting of billions of examples. Nevertheless, these methods did not consider the integration of the text feature and image feature for image generation, they indeed did not perform well in  LPIPS and FID.

Table~\ref{tab:time} compares the model memory and average time cost for synthesizing an image of size 256 $\times$ 256 on a RTX 3090 GPU. As shown, the running time of the SGDiff model is much shorter than that of VQGAN-CLIP and FuseDream. Although the running time of the SGDiff model is slightly longer than LDM, the memory consumption is lower. Compared to the baseline, the increases in time and memory are relatively insignificant because we only fine tune the image encoder and modality fusion module. In summary, the SGDiff can be trained without much memory and can generate an image with good quality based on text and style conditions within 10 seconds on RTX 3090.

\begin{table}[t]
 \begin{threeparttable}
\centering
    \caption{Ablation experiments on modality fusion methods and classifier-free approaches.}
    \vspace{-5px}
    \label{tab:setting ablation}
     \begin{minipage}{\textwidth}
    \begin{tabular}{cccc@{\hspace{0.5em}}c@{\hspace{0.5em}}c@{\hspace{0.5em}}}
    \toprule Classifier-free & Mask & Modality fusion & LPIPS $\downarrow$ & FID $\downarrow$ & CS $\uparrow$ \\
    \cmidrule(lr){1-3}
    \cmidrule(lr){4-6}
   Eq. (\ref{eq:classifier-free})  &  &$\oplus$\footnotemark[1] & 0.6833 & 42.63 & 25.63 \\    
   Eq. (\ref{eq:classifier-free})  &  & CA\footnotemark[2]& 0.5650 & 38.88 & 25.39 \\
    Eq. (\ref{eq:classifier-free})  &  & SCA & 0.5607 & 39.21 & 25.98 \\
    Eq. (\ref{eq:classifier-free})  & \checkmark & SCA & 0.5695 & 37.22 & 26.06 \\
    Eq. (\ref{eq:multi-classifier-free}) & \checkmark & SCA & \textbf{0.4474} & \textbf{32.06} & \textbf{27.53} \\
    \bottomrule
    \end{tabular}
    \end{minipage}
    \begin{tablenotes}
    \footnotesize
        \item [1] $\oplus$ refers to an element-wise addition operation, where the features $f_T$ and $f_S$ are projected onto the same dimension before opeation; 
        \item [2] CA indicates SCA module without skip connection, w.r.t. Eq.~(\ref{eq:CA}) withouth Eq.~(\ref{eq:skip}). 
   \end{tablenotes}
   \vspace{-10px}
\end{threeparttable}    
\end{table}

\subsection{Ablation Study}
\label{sec:ablation}
Ablation study was conducted to evaluate the effect of each component of the proposed SGDiff on SG-Fashion dataset.

\textbf{Effectiveness of the SCA:} 
As demonstrated in Table~\ref{tab:setting ablation}, the comparison between the element-wise addition of features and the cross-attention (CA) method shows that CA is significantly more effective in improving LPIPS and FID scores. However, it has the downside of causing a decline in semantic information, as CS decreases. To address this issue, the SCA moduel with skip connections was use. As shown in the third row of the table, SCA leads to improvements in both LPIPS and CS scores, demonstrating its ability to improve the similarity between synthesized images and ground truth images.

\textbf{Effectiveness of background masking:} 
As shown in Table~\ref{tab:setting ablation}, after applying background masking, the FID value decreases by 1.99 and the CS remains almost the same. This demonstrates that background masking is beneficial to improve image quality. The reason for slightly increased LPIPS is that LPIPS is sensitive to perceptual information, the lack of background may degrade LPIPS metric. However, the fashion synthesis task only focuses on the synthesized foreground, and the background could be easily removed.

\textbf{The orders and weights for different conditions:}  
Figure~\ref{fig:ablation} displays the relationship between FID, LPIPS and CS with different conditional weights and order settings. The trend of setting text prior to style is similar to setting style prior to text, indicating little impact on results with fixed $s_S=1$ and varying $s_T$. In addition, it can be seen from Figure~\ref{fig:ablation} that the optimal values (see the circled dots of Figure~\ref{fig:ablation}) of $s_S$ and $s_T$ are almost in the range of 1.0 to 1.6. More specifically, we choose the setting of $s_S=1.2$, $s_T=1.0$, and \emph{style prior to text} as optimal. This setting achieves the best LPIPS which is important in controlling synthesized styles. The numerical results are shown in the last row of Table~\ref{tab:setting ablation} 

\section{Conclusions and Future Work}
This paper has reported on the development of a novel style guided diffusion model (SGDiff), overcoming inherent weaknesses in existing diffusion models for image synthesis.  
The proposed SGDiff has demonstrated its effectiveness in incorporating style guidance into pretrained text-to-image diffusion models. Without relying on large amounts of labelled data or computing resources, SGDiff is capable of achieving promising control over the synthesized textures, making it a valuable contribution to the field. 
As a future work, we plan to expand upon the capabilities of SGDiff by focusing on more detailed control over various aspects of the synthesized textures, such as color themes, patterns, and materials. By refining these controls, we aim to further improve the utility and applicability of the proposed model in diverse applications and domains.

\begin{acks}
The work described in this paper is supported in part by the Innovation and Technology Commission of Hong Kong under grant ITP/028/21TP and by the Laboratory for Artificial Intelligence in Design (Project Code: RP1-1) under InnoHK Research Clusters, Hong Kong Special Administrative Region.
\end{acks}

\bibliographystyle{ACM-Reference-Format}
\balance
\bibliography{main}

\end{document}